\documentclass{article}
\usepackage{spconf,amsmath,graphicx,hyperref}
\usepackage{amssymb}
\usepackage{comment}

\title{Parts-Mamba: Augmenting Joint Context with Part-Level Scanning for Occluded Human Skeleton}
\name{Tianyi Shen\textsuperscript{1}, Huijuan Xu\textsuperscript{1}, Nilesh Ahuja\textsuperscript{2}, Omesh Tickoo\textsuperscript{2}, Philip Shin\textsuperscript{1}, Vijaykrishnan Narayanan\textsuperscript{1}\thanks{This work was supported in part by SRC/DARPA JUMP 2.0 PRISM Center.}}
\address{\textsuperscript{1}The Pennsylvania State University, State College, PA, United States, \\ \textsuperscript{2}Intel Corp., Santa Clara, CA, United States}
\begin{document}
%
\maketitle
\begin{abstract}
Skeleton action recognition involves recognizing human action from human skeletons. The use of graph convolutional networks (GCNs) has driven major advances in this recognition task. In real-world scenarios, the captured skeletons are not always perfect or complete because of occlusions of parts of the human body or poor communication quality, leading to missing parts in skeletons or videos with missing frames. In the presence of such non-idealities, existing GCN models perform poorly due to missing local context. To address this limitation, we propose Parts-Mamba, a hybrid GCN-Mamba model designed to enhance the ability to capture and maintain contextual information from distant joints. The proposed Parts-Mamba model effectively captures part-specific information through its parts-specific scanning feature and preserves non-neighboring joint context via a parts-body fusion module. 
Our proposed model is evaluated on the NTU RGB+D 60 and NTU RGB+D 120 datasets under different occlusion settings, achieving up to 12.9\% improvement in accuracy.
\end{abstract}
\begin{keywords}
state space model, skeleton-based human action recognition, mamba, occlusion, video understanding
\end{keywords}
\begin{figure*}[ht]
    \centering
    \centerline{\includegraphics[width=0.7\textwidth]{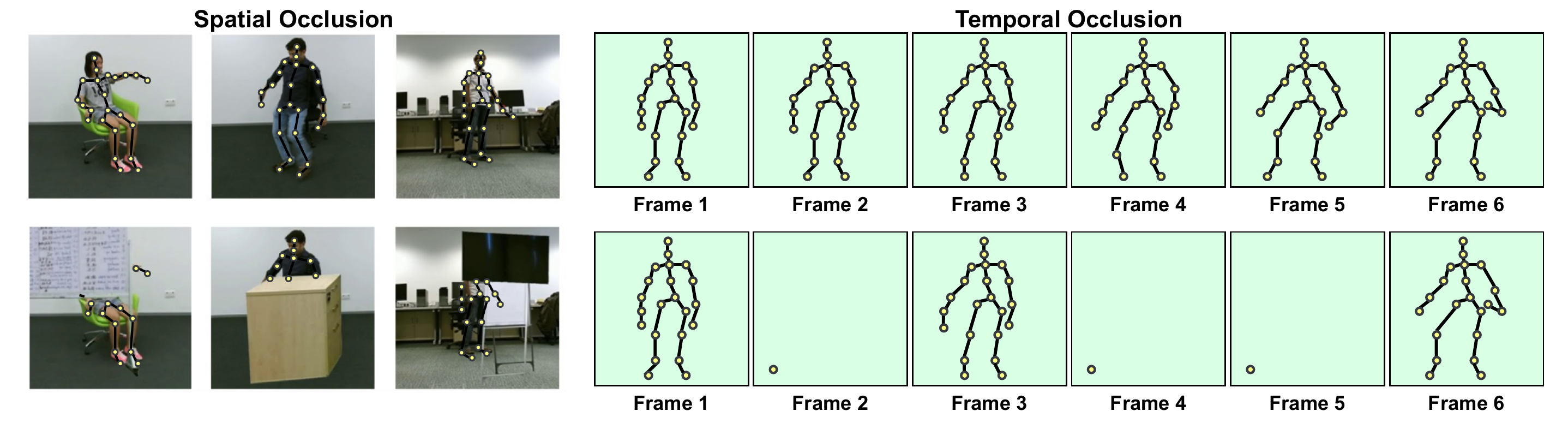}}
    \vspace{-0.4cm}
    \caption{An illustration of spatial occlusion and temporal occlusion. The figures at the top show the skeleton detected from regular video input, and the bottom figures show the occluded skeleton detected from the video input with occlusion. Note that in temporal occlusion, all the nodes are collapsed to one original node in occluded frames.}
    \vspace{-0.4cm}
    \label{fig:occluded_scene}
\end{figure*}
\section{Introduction}
Human action recognition plays a critical role in various domains, including human-computer interaction, robotics, home security, and surveillance. Compared to RGB and depth data, skeleton data usually contains a more concise information representation and facilitates capturing joint synergies in the human body.

Numerous efforts attempt to enhance classification accuracy for skeleton-based action recognition~\cite{stgcn,skeleton_mixformer,CTR-GCN,HDGCN}. GCN dominates the field due to its strong ability to capture the spatial and temporal relationships among human joints. While existing state-of-the-art models have shown excellent performance on non-occluded skeleton data, real-world scenarios often involve occlusions in both skeletons and frames. As shown in Fig.~\ref{fig:occluded_scene}, when an object blocks the view between the human body and the monitoring camera, the extracted skeleton can become corrupted, leading to missing body parts. Additionally, unstable connections during signal transmission can result in blank frames with zero values for all skeleton joints. In those cases, conventional graph convolutional networks (GCNs) fail to maintain their high performance on occluded skeletons and frames. A potential reason is that GCN primarily focuses on neighboring nodes and frames, which makes it less effective at capturing contextual information from distant joints. When occlusion happens, it tends to disrupt the context of neighboring joints or frames, which can be fatal for GCN to maintain its performance on lossless skeleton video data.

Most existing methods that handle occluded skeletal data~\cite{RA-GCN,PDGCN} concentrate on improving local and part-wise modeling by employing multi-stream GCN architectures where each stream captures part-specific information. However, they often overlook the critical importance of distant joint relationships that play roles when a critical frame is lost or a body part is missing. Therefore, adding additional modules designed for global context extraction is necessary.

As a recently proposed model that emphasizes both efficient long-sequence processing and robust long-range joint dependency modeling, Mamba becomes an excellent candidate for enhancing GCN-based approaches in preserving the contextual information of distant joints. 

\textbf{Contributions.} With the unique scanning mechanism of Mamba, we introduce PartsMamba, which performs selective joint scanning for parts-specific modeling while preserving global body information through a dedicated Part-Wise and Body Scanning module. We further incorporate topological graph modeling to embed domain-specific graph knowledge and learn adaptive graph topologies. All these representations are subsequently fused via a gated Mamba fusion module, yielding the final outputs of the Mamba spatial encoder. Moreover, the Mamba temporal encoder processes input sequences along the temporal dimension and applies spatial-temporal scanning to capture complex spatial-temporal correlations. These design choices significantly enhance the model’s ability to capture and maintain distant spatial context under occlusion scenarios.
\vspace{-0.4cm}
\section{Details of PartsMamba}
In our proposed PartsMamba architecture, we choose the DeGCN~\cite{degcn}, which is the current SOTA model, as the GCN part of the hybrid structure to dynamically select the joints in the skeleton during GCN feature extraction. The PartsMamba framework includes a series of Spatial-Temporal blocks that contain a Mamba spatial fusion block and a Mamba temporal block to help capture and maintain the spatial-temporal information of the processed skeleton sequence.

\vspace{-0.4cm}
\subsection{Part-Wise Scanning and Body Scanning}
\begin{figure}[ht]
    \centering
    \centerline{\includegraphics[width=0.5\textwidth]{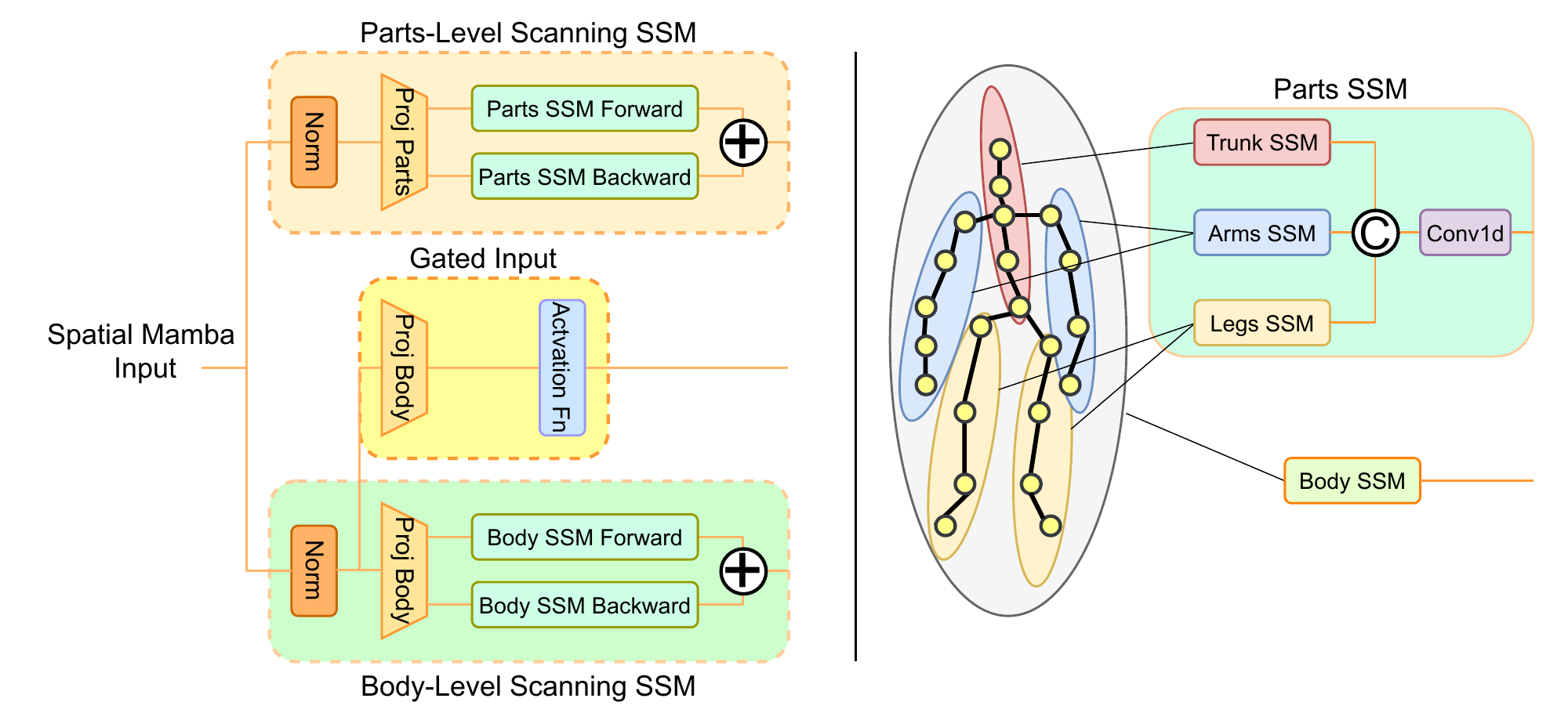}}
    \caption{Detailed structures of the parts-body scanning module.}
    \label{fig:parts_scanning}
\end{figure}
To tackle the possible occlusion scenario and improve the robustness of the model, we adopt the idea of LocalMamba~\cite{huang2024localmamba} that processes scanning on grouped image patches to enhance the model's local feature extraction ability. In our proposed parts-wise scanning module shown in Fig.\ref{fig:parts_scanning}, the processed input skeleton sequence \(X_{gcn}\in\mathbb{R}^{V\times T\times C}\) where \(T\) is the temporal window size sampled from the raw video frames, \(V\) is the number of skeleton joints, and \(C\) is the output channel dimension of the GCN head will first be normalized with layer normalization and then projected to a smaller inner dimension as follows:
\begin{equation}
\begin{matrix}
    X_{p}=LayerNorm(W_{p}X_{gcn}),\\
    \\
    X_{s}=LayerNorm(W_{s}X_{gcn}),\\
    \\
    X_{g}=LayerNorm(W_{s}X_{gcn}).
\end{matrix}
\end{equation}
where the down-projected \(X_{p}\in\mathbb{R}^{V\times T\times C^{'}}\) (\(C^{'}\) is the down-projected channel dimension) will be the input of the Parts-Level Scanning SSM \((SSM_p)\) and the normalized and dowe-projected \(X_{s}\in\mathbb{R}^{V\times T\times C^{'}}\) will be the input for both the Gated Input and the Body-Level Scanning SSM \((SSM_s)\) (Fig. \ref{fig:parts_scanning}). The Parts-Level Scanning will process the input through different parts across the entire body, and the Body-Level Scanning directly processes through the whole skeleton to form a local-feature-intensive representation and a global-feature-intensive representation. Along with the gated value representing the original skeleton information, the whole process can be described as follows:
\begin{equation}
\begin{matrix}
    X_{p}=Conv1D(\parallel_{\rho \in P}SSM_{\rho}(X_{p})\oplus SSM_{\rho}(X_{p}^{r})),\\
    \\
    X_{s}=SSM_{s}(X_{s})\oplus SSM_{s}(X_{s}^{r}).
\end{matrix}
\end{equation}
\begin{equation}
    X_{g}=ReLU(X_{g}).
\end{equation}
where \(\rho\in\mathbb{R}^{V^{'}}\) \((V^{'}\) represents the size of the selected joints subsets) is the parts subset selected from the whole body \(P\in\mathbb{R}^{V}\) and \(SSM_{\rho}\) denotes the independent SSM modules receiving different body parts so each SSM can learn a unique feature representation exclusively for the designated parts subset. \(\oplus\) stands for the element-wise plus, and \(\parallel\) stands for the channel concatenation of the output tensor from different parts SSMs. Input with the reversed joints order \(X_{p}^{r}, X_{s}^{r}\in\mathbb{R}^{V\times T\times C^{'}}\) will be further added to the forward result to form the bidirectional architecture that helps to better capture the correlations between neighboring nodes, and a conv1d is used after the parts SSM to mix the features from different channels. Then, the gated unit \(X_g\) is obtained with the ReLU activation function.

\vspace{-0.4cm}
\subsection{Mamba with Topological Graph Modeling}
The computed SSM results cannot be used directly because they only contain the joint features without any graph knowledge added. Inspired by the approach from CTR-GCN~\cite{CTR-GCN}, we incorporated the graph information into the SSM results as follows:
\begin{equation}
\begin{matrix}
    X_{p}^{'}=LayerNorm(\Lambda(\alpha X_{p}, A_{p})),\\
    \\
    X_{s}^{'}=LayerNorm(\Lambda(\alpha X_{s}, A_{s})).
\end{matrix}
\end{equation}
where \(\Lambda\) is the channel aggregation function, and \(A_p, A_s\in\mathbb{R}^{V\times V}\) are the learnable shared topologies incorporated with graph knowledge. We put a trainable scalar \(\alpha\in\mathbb{R}\) to adjust the impact ratio from SSM output features. The resulting \(X_{p}^{'}\in\mathbb{R}^{V\times T\times C^{'}}\) and \(X_{s}^{'}\in\mathbb{R}^{V\times T\times C^{'}}\) will be used as graph-aware inputs in the Gated Mamba Fusion module. 

\vspace{-0.4cm}
\subsection{The Gated Mamba Fusion Module}
At this point, there will be five inputs generated from all the previous modules in the Mamba Spatial Fusion block: \(X_{p}\) and \(X_{s}\) represent the SSM outputs of joint-level context for neighboring nodes and distant nodes with no graph knowledge added; \(X_{p}^{'}\) and \(X_{s}^{'}\) denote the SSM results of graph-level context for neighboring nodes and distant nodes; \(X_{g}\) is the residual input feature that helps reference encoder input. The module fuses all those tensors in the following way:
\begin{equation}
\begin{matrix}
    F_{self} =(X_{p}*X_{g})\oplus (X_{s}*X_{g}),\\
    \\
    F_{cross} =(X_{p}^{'}*X_{s})\oplus (X_{p}*X_{s}^{'}),\\
    \\
    F = W_{f}(F_{self}\oplus F_{cross}).
\end{matrix}
\end{equation}
where \(*\) is the element-wise multiplication. The self-fusion involves the star operation result between \(X_{p}, X_{g}\) that combines the processed parts information with the original skelton information and between \(X_{s}, X_{g}\) that combines the global skeleton information captured by SSM with the original skeleton input. \(F_{self}\in\mathbb{R}^{V\times T\times C^{'}}\) comes from adding up the result of parts information \(X_{p}*X_{g}\) and the skeleton information \(X_{s}*X_{g}\). The cross-fusion result \(F_{cross}\in\mathbb{R}^{V\times T\times C^{'}}\) comes from adding cross-graph relations \(X_{p}^{'}*X_{s}\) that gives the graph-infused result with focus on parts, and \(X_{p}*X_{s}^{'}\) that gives the graph-infused result with focus on body. The self-fusion result and the cross-fusion result will be added up to give the combined feature and up-projected to form the final fusion result \(F\in\mathbb{R}^{V\times T\times C}\).

\vspace{-0.4cm}
\subsection{Mamba Temporal Encoder}
The Mamba temporal encoder is designed to model the long-range dependencies and capture temporal dynamics presented across consecutive frames in skeleton-based action sequences. As the graph becomes several non-connected 1-D movement lines in the temporal domain, excessive design for such simple graphs is unnecessary. So, as the Mamba encoder itself is already good enough for sequential modeling, we use the standard Mamba encoder with a unidirectional selective scanning mechanism as the basic building block with the input swapping dimension between \(V\) and \(T\): \(X\in\mathbb{R}^{T\times V\times C}\).\\
\textbf{Spatial-Temporal Scanning on Skeleton Data.} To further enhance the model performance of modeling the spatial-temporal relations. We adopted the idea from~\cite{videomamba}, which adds both in-frame and cross-frame contents into the scanning path for video understanding and uses it on the skeleton data. Given the input \(X\in\mathbb{R}^{T\times V\times C}\), instead of scanning frame by frame with dimension \(V*C\), our proposed spatial-temporal scanning split the scanning of a single frame into several sub-frame sections with dimension \((V/N)*C\), where \(N\) is the number of sub-frame sections split, and add them into the scanning path to form \(X\in\mathbb{R}^{(T*N)\times (V/N)\times C}\). 
\section{Experiments}
\vspace{-0.4cm}
\begin{table}[ht]
\begin{center}
\resizebox{\columnwidth}{!}{
\begin{tabular}{l|ccccccc}
\hline
\textbf{} &\multicolumn{6}{c}{\textbf{NTU-60 Cross-Subject}}\\
\textbf{Model}&\multicolumn{6}{c}{\textbf{Occluded Part}}\\
\hline
&\textbf{None}&\textbf{Left Arm}&\textbf{Right Arm}&\textbf{Two Hands}&\textbf{Two Legs}&\textbf{Trunk}&\textbf{Mean}\\
\hline 

\textbf{ST-GCN}                     & 80.7 & 71.4 & 60.5 & 62.6 & 77.4 & 50.2 & 64.4\\
\textbf{SR-TSL}                     & 84.8 & 70.6 & 54.3 & 48.6 & 74.3 & 56.2 & 60.8\\
\textbf{2s-AGCN}                    & 88.5 & 72.4 & 55.8 & \textbf{82.1} & 74.1 & 71.9 & 71.3\\
\textbf{1s RA-GCN}                  & 85.8 & 69.9 & 54.0 & 66.8 & 82.4 & 64.9 & 67.6\\
\textbf{2s RA-GCN}                  & 86.7 & 75.9 & 62.1 & 69.2 & 83.3 & 72.8 & 72.7\\
\textbf{3s RA-GCN}                  & 87.3 & 74.5 & 59.4 & 74.2 & 83.2 & 72.3 & 72.2\\
\textbf{STIGCN}                     & 88.8 & 12.7 & 11.5 & 18.3 & 45.5 & 20.9 & 21.8\\
\textbf{MS-G3D}                     & 87.3 & 31.3 & 23.8 & 17.1 & 78.3 & 61.6 & 42.4\\
\textbf{CTR-GCN}                    & 87.5 & 13.0 & 12.5 & 12.7 & 21.0 & 36.3 & 19.1\\
\textbf{TCA-GCN}                    & 90.2 & 75.4 & 53.4 & 70.8 & 75.2 & 78.6 & 70.7\\
\textbf{HD-GCN}                     & 86.8 & 67.1 & 55.7 & 56.7 & 74.8 & 61.3 & 63.1\\
\textbf{4s MST-GCN}                 & \textbf{91.5} & 73.8 & 57.9 & 77.1 & 82.8 & 82.4 & 74.8\\
\textbf{1s PD-GCN}                  & 86.8 & 67.1 & 55.7 & 56.7 & 74.8 & 61.3 & 63.1\\
\textbf{2s PD-GCN}                  & 87.4 & 76.4 & 62.0 & 74.4 & 84.8 & 70.4 & 73.6\\
\textbf{3s PD-GCN}                  & 87.5 & 76.0 & 62.0 & 75.4 & 85.0 & 73.0 & 74.3\\
\textbf{MSFGCN}                     & 87.2 & 82.3 & 73.6 & 74.0 & 86.0 & \textbf{85.4} & 80.3\\
\hline 
\textbf{PartsMamba (ours)}         & 90.8 & \textbf{87.3} & \textbf{82.7} & 81.4 & \textbf{88.3} & 82.3 & \textbf{84.4}\\

\hline
\hline
\end{tabular}
}
\caption{Parts-Occlusion Results on NTU-60 Dataset.}
\label{tab:parts_60}
\end{center}
\end{table}
\vspace{-0.8cm}
\begin{table}[ht]
\begin{center}
\resizebox{\columnwidth}{!}{
\begin{tabular}{l|ccccccc}
\hline
\textbf{} &\multicolumn{6}{c}{\textbf{NTU-120 Cross-Setup}}\\
\textbf{Model}&\multicolumn{6}{c}{\textbf{Occluded Part}}\\
\hline
&\textbf{None}&\textbf{Left Arm}&\textbf{Right Arm}&\textbf{Two Hands}&\textbf{Two Legs}&\textbf{Trunk}&\textbf{Mean}\\
\hline 

\textbf{ST-GCN}                     & 73.2 & 59.7 & 47.3 & 52.5 & 68.5 & 48.5 & 55.3\\
\textbf{SR-TSL}                     & 79.9 & 59.4 & 50.3 & 41.2 & 64.8 & 55.0 & 54.1\\
\textbf{2s-AGCN}                    & 84.2 & 62.8 & 46.6 & 77.8 & 67.0 & 60.7 & 63.0\\
\textbf{1s RA-GCN}                  & 80.0 & 64.0 & 49.7 & 50.0 & 74.7 & 60.2 & 59.7\\
\textbf{2s RA-GCN}                  & 82.5 & 64.7 & 54.1 & 56.0 & 77.6 & 67.7 & 64.0\\
\textbf{3s RA-GCN}                  & 82.7 & 68.5 & 54.9 & 57.5 & 79.0 & 69.9 & 66.0\\
\textbf{4s MST-GCN}                 & 88.8 & 68.5 & 55.1 & 66.3 & 78.7 & 79.0 & 69.5\\
\textbf{MSFGCN}                     & 81.4 & 75.0 & 66.8 & 63.6 & \textbf{81.7} & 80.1 & 73.4\\
\hline 
\textbf{PartsMamba (ours)}         & \textbf{90.2} & \textbf{84.8} & \textbf{79.0} & \textbf{87.7} & 79.4 & \textbf{82.6} & \textbf{82.7}\\

\hline
\hline
\end{tabular}
}
\caption{Parts-Occlusion Results on NTU-120 Dataset.}
\label{tab:parts_120}
\end{center}
\end{table}
\vspace{-1cm}
\subsection{Experiment Setup}
\textbf{Dataset.} To test the effectiveness of the proposed architecture, we followed the procedure from previous works to test our model on the NTU-60~\cite{ntu60} and NTU-120~\cite{ntu120} datasets. Following the settings proposed by~\cite{RA-GCN}, the spatial occlusion was evaluated under the cross-subject setting, and the temporal occlusion was evaluated under the cross-setup setting.\\
\textbf{Evaluation Metrics.} We choose two evaluation settings from the previous works: \textbf{parts-specific occlusion} and \textbf{temporal occlusion}. The parts-specific occlusion is used to model the scenario in which there is an object between the camera and the human body, causing a missing part in the captured video. In that setting, five occlusion scenes are used for model evaluation: left arm occlusion, right arm occlusion, two hands occlusion, two legs occlusion, and trunk occlusion. The temporal occlusion is used to model a scenario in which the camera encounters a sudden signal loss during data transmission to the server. In that setting, continuous frames with specific portions will be masked from a random starting point.

\textbf{Training and Evaluation.} To ensure the graph convolution part and the Mamba encoder do not affect the learning of each other, we performed a two-step training that trained the graph convolution part first and used its pre-trained weights for the training of the hybrid model. 

For the parts-specific occlusion setting, five occlusion scenes plus the non-occluded case will be randomly chosen, and the input training data will be multiplied by a mask to help the model learn the occluded feature representation. In this setting, 240 epochs were used for the training of the graph convolution heads, and another 240 epochs were used for the training of the hybrid model. For the temporal frame occlusion settings, the model was trained with the non-occluded data and tested on the occluded data. In those settings, 80 epochs were used for the training of graph convolution heads and 140 were used for the hybrid model training. All the training was performed on a single A100 for mixed precision training with a batch size of 64. 


\vspace{-0.4cm}
\subsection{Results on Parts-Specific Occlusion}
The performance of our model under parts-specific occlusion is reported in Tables \ref{tab:parts_60} and \ref{tab:parts_120}. Compared to previous approaches, our method achieves state-of-the-art performance, giving a mean accuracy of 84.4\% on the NTU-60 dataset and 82.7\% on the NTU-120 dataset, outperforming the strongest existing model, MSFGCN~\cite{MSFGCN}, by 4.1\% and 9.3\%, respectively. Notably, the primary improvements come from the “left arm occlusion” and “right arm occlusion” scenarios in both datasets, as well as the “two hands occlusion” in NTU-120, indicating that our model is especially robust for upper-body occlusions.

\begin{table}[ht]
\begin{center}
\resizebox{\columnwidth}{!}{
\begin{tabular}{l|ccccccc}
\hline
\textbf{} &\multicolumn{6}{c}{\textbf{NTU-60 Cross-Subject}}\\
\textbf{Model}&\multicolumn{6}{c}{\textbf{Portion of Masked Frames}}\\
\hline
&\textbf{0\%}&\textbf{10\%}&\textbf{20\%}&\textbf{30\%}&\textbf{40\%}&\textbf{50\%}&\textbf{Mean}\\
\hline 

\textbf{ST-GCN}                     & 80.7 & 69.3 & 57.0 & 44.5 & 34.5 & 24.0 & 45.9\\
\textbf{SR-TSL}                     & 84.8 & 70.9 & 62.6 & 48.8 & 41.3 & 28.8 & 50.5\\
\textbf{2s-AGCN}                    & 88.5 & 74.8 & 60.8 & 49.7 & 38.2 & 28.0 & 50.3\\
\textbf{1s RA-GCN}                  & 85.8 & 81.6 & 72.9 & 61.6 & 47.9 & 34.0 & 59.6\\
\textbf{2s RA-GCN}                  & 86.7 & 83.0 & 76.4 & 65.6 & 53.1 & 39.5 & 63.5\\
\textbf{3s RA-GCN}                  & 87.3 & 83.9 & 76.4 & 66.3 & 53.2 & 38.5 & 63.7\\
\textbf{STIGCN}                     & 88.8 & 70.4 & 51.0 & 38.7 & 23.8 & 8.0 & 38.4\\
\textbf{MS-G3D}                     & 87.3 & 77.6 & 65.7 & 54.3 & 41.9 & 30.1 & 53.9\\
\textbf{CTR-GCN}                    & 87.5 & 72.4 & 54.1 & 35.6 & 22.4 & 11.5 & 39.2\\
\textbf{TCA-GCN}                    & 90.2 & 84.4 & 74.6 & 58.1 & 42.3 & 25.6 & 57.0\\
\textbf{HD-GCN}                     & 86.8 & 57.0 & 29.5 & 18.5 & 11.2 & 7.04 & 24.7\\
\textbf{1s PD-GCN}                  & 85.7 & 81.9 & 75.4 & 66.4 & 54.9 & 40.0 & 63.7\\
\textbf{2s PD-GCN}                  & 87.4 & 83.8 & 76.7 & 66.8 & 55.1 & 40.6 & 64.6\\
\textbf{3s PD-GCN}                  & 87.5 & 83.9 & 76.6 & 66.7 & 53.9 & 40.0 & 64.2\\
\hline 
\hline 
\textbf{PartsMamba(ours)}       & \textbf{92.4} & \textbf{90.6} & \textbf{87.7} & \textbf{80.9} & \textbf{71.1} & \textbf{53.6} & \textbf{76.8}\\

\hline
\hline
\end{tabular}
}
\caption{Temporal Occlusion Results on NTU-60 Dataset.}
\label{tab:cont_rand_60}
\end{center}
\end{table}
\vspace{-0.4cm}
\vspace{-0.4cm}
\begin{table}[h]
\begin{center}
\resizebox{\columnwidth}{!}{
\begin{tabular}{l|ccccccc}
\hline
\textbf{} &\multicolumn{6}{c}{\textbf{NTU-120 Cross-Setup}}\\
\textbf{Model}&\multicolumn{6}{c}{\textbf{Portion of Masked Frames}}\\
\hline
&\textbf{0\%}&\textbf{10\%}&\textbf{20\%}&\textbf{30\%}&\textbf{40\%}&\textbf{50\%}&\textbf{Mean}\\
\hline 

\textbf{ST-GCN}                     & 73.2 & 60.8 & 48.8 & 38.2 & 27.3 & 17.4 & 38.5\\
\textbf{SR-TSL}                     & 79.9 & 67.4 & 58.8 & 50.4 & 44.7 & 37.1 & 51.7\\
\textbf{2s-AGCN}                    & 84.2 & 73.4 & 56.8 & 44.2 & 33.0 & 23.4 & 46.2\\
\textbf{1s RA-GCN}                  & 80.0 & 76.1 & 67.9 & 57.1 & 44.0 & 30.3 & 55.1\\
\textbf{2s RA-GCN}                  & 82.5 & 79.1 & 72.4 & 63.3 & 51.2 & 36.6 & 60.5\\
\textbf{3s RA-GCN}                  & 82.7 & 79.6 & 72.9 & 63.3 & 51.2 & 36.8 & 60.8\\
\hline 
\textbf{PartsMamba - middle}       & \textbf{90.2} & \textbf{87.9} & \textbf{84.8} & \textbf{77.6} & \textbf{67.3} & \textbf{50.9} & \textbf{73.7}\\
\hline 
\textbf{PartsMamba - random}       & 90.2 & 87.9 & 84.9 & 78.1 & 68.1 & 51.6 & 74.1\\

\hline
\hline
\end{tabular}
}
\caption{Temporal Occlusion Results on NTU-120 Dataset.}
\vspace{-0.8cm}
\label{tab:cont_rand_120}
\end{center}
\end{table}
\vspace{-0.4cm}
\subsection{Results on Temporal Frame Occlusion}
We evaluate our model under temporal frame occlusion settings in Tables \ref{tab:cont_rand_60} and \ref{tab:cont_rand_120}. 
In these settings, we randomly select an initial point (ensuring sufficient remaining frames for masking) and obscure the subsequent frames. Since there exists randomness in random occlusion, we repeat the procedure five times and report the mean accuracy across all the experiments. Under these conditions, our model achieves state-of-the-art performance on both NTU-60 and NTU-120, with improvements of 10.8\% and 12.9\% of the averaged occlusion accuracy over the previous best models.

\vspace{-0.4cm}
\subsection{Case Study with Other Architectures}
\vspace{-0.4cm}
\begin{figure}[ht]
    \centering
    \centerline{\includegraphics[width=0.5\textwidth]{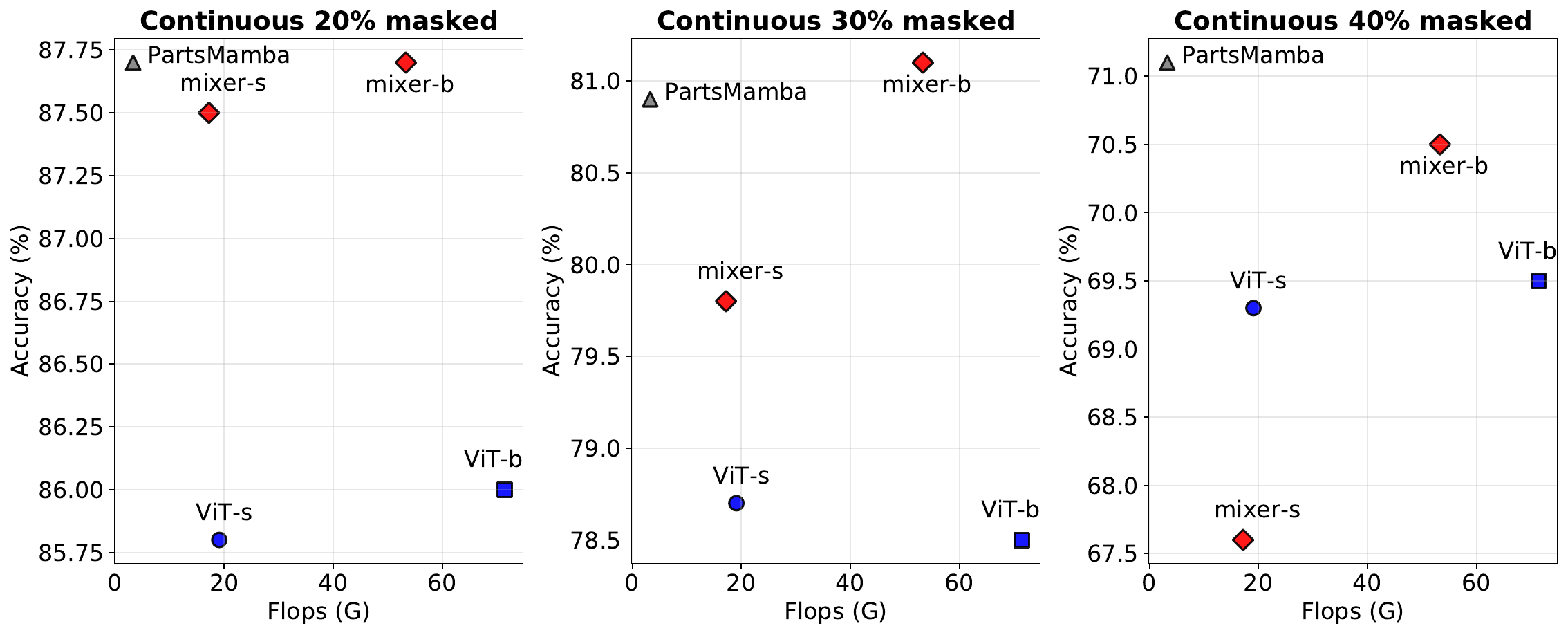}}
    \caption{Performance Analysis After Replacing the PartsMamba Block by Vision Transfomer and MLPMixer.}
    \label{fig:manmba_flops}
\end{figure}
\textbf{Performance and Efficiency Analysis.}
We further evaluated the proposed Mamba architecture by replacing its context extractor with the Vision Transformer (ViT)~\cite{vit} or MLPMixer~\cite{mixer} while keeping all other training parameters identical to PartsMamba. We trained and tested ViT-small, ViT-base, MLPMixer-small, and MLPMixer-base on the temporal occlusion scenarios, recording both floating-point operations (FLOPS) and inference speed (FPS), as illustrated in Fig.~\ref{fig:manmba_flops}.

In all three occlusion settings, PartsMamba outperforms the ViT's and MLPMixer's baseline accuracy (y-axis in Figure.~\ref{fig:manmba_flops}) in the 20\% masked case and 40\% masked case. Moreover, the computational overhead of PartsMamba (3.3G FLOPS) is substantially lower than that of ViT-small (19.1G FLOPS), MLPMixer-small (17.2G FLOPS), ViT-base (70.4G FLOPS), and MLPMixer-base (53.3G FLOPS). 
\vspace{-0.4cm}
\section{Conclusion}
In this work, we present a novel Mamba-based hybrid architecture, PartsMamba, for occluded skeleton-based action recognition. Our experiment and case study results demonstrate its great ability to model distant joint dependencies under occlusion settings. On all four occlusion settings, the proposed model achieves state-of-the-art performance out of all the previous works. Compared with using other context extractor modules, the proposed Mamba block shows great FLOPs reduction while still achieving the best performance.

\bibliographystyle{IEEEbib}
\bibliography{strings,refs}

\end{document}